%% file: pkdd2015-top-interval-patterns.tex
\documentclass{article}
\usepackage{hyperref}

\input{tex/packages}

\newcommand{\inst}[1]{$^{#1}$}
\newcommand{\keywords}[1]{\par\noindent{\bf KEYWORDS:} #1}
\usepackage{amsmath}
\newtheorem{definition}{Definition}
\newtheorem{proposition}{Proposition}
\newtheorem{example}{Example}

\input{tex/settings}

\input{tex/commands}

\input{tex/graphicks.tikz.tex}

\begin{document}

\title{
	Fast Generation of Best Interval Patterns for Nonmonotonic Constraints\footnote{The final publication is available at \href{http://link.springer.com}{link.springer.com}}
}
\author{
	Aleksey Buzmakov\inst{1,2} \and Sergei O. Kuznetsov\inst{2} \and Amedeo Napoli\inst{1}\\
}
\date{
	\inst{1}LORIA (CNRS -- Inria NGE -- U. de Lorraine), Vandœuvre-lès-Nancy, France
	\\ 
	\inst{2}National Research University Higher School of Economics, Moscow, Russia
	\\
	aleksey.buzmakov@inria.fr, skuznetsov@hse.ru, amedeo.napoli@loria.fr
}

%
\maketitle
%
\setcounter{footnote}{0}

\begin{abstract}
	In pattern mining, the main challenge is the exponential explosion of the set of patterns. Typically, to solve this problem, a constraint for pattern selection is introduced. One of the first constraints proposed in pattern mining is support (frequency) of a pattern in a dataset. Frequency is an anti-monotonic function, i.e., given an infrequent pattern, all its superpatterns are not frequent. However, many other constraints for pattern selection are not (anti-)monotonic, which makes it difficult to generate patterns satisfying these constraints.
	In this paper we introduce the notion of projection-antimonotonicity and \ThetaAlgoName{} algorithm that allows efficient generation of the best patterns for some nonmonotonic constraints.
	In this paper we consider stability and \textDelta-measure, which are nonmonotonic constraints, and apply them to interval tuple datasets. In the experiments, we compute best interval tuple patterns w.r.t. these measures and show the advantage of our approach over postfiltering approaches.
\end{abstract}

\section{Introduction}
Interestingness measures were proposed to overcome the problem of combinatorial explosion of the number of valid patterns that can be discovered in a dataset~\cite{Vreeken2014}.
For example, pattern support, i.e., the number of objects covered by the pattern, is one of the most famous measures of pattern quality.
In particular, support satisfies the property of anti-monotonicity (aka ``a priori principle''), i.e., the larger the pattern is the smaller the support is~\cite{Mannila1994,Agrawal1994}.
Many other measures can be mentioned such as utility constraint~\cite{Yao2006}, pattern stability~\cite{KuznetsovStability2007,Roth2008}, pattern leverage~\cite{Webb2010}, margin closeness~\cite{Moerchen2011}, MCCS~\cite{Spyropoulou2013}, cosine interest~\cite{Cao2014}, pattern robustness~\cite{Tatti2014}, etc.

Some of these measures (e.g., support, robustness for generators~\cite{Tatti2014}, or upper bound constraint of MCCS~\cite{Spyropoulou2013}) are ``globally anti-monotonic'', i.e., for any two patterns $X \sqsubseteq Y$ we have $\M(X) \geq \M(Y)$, where $\M$ is a measure and $\sqsubseteq$ denotes the (subsumption) order relation on patterns.
When a measure is anti-monotonic, it is relatively easy to find patterns whose measure is higher than a certain threshold (e.g., patterns with a support higher than a threshold).
In contrast some other measures are called ``locally anti-monotonic'', i.e., for any pattern $X$ there is an immediate subpattern $Y \prec X$ such that $\M(Y) \geq \M(X)$.
Then the right strategy should be selected for traversing the search space, e.g., a pattern $Y$ should be extended only to patterns $X$ such that $\M(Y) \geq \M(X)$.
For example, for ``locally anti-monotonic'' cosine interest~\cite{Cao2014}, the extension of a pattern $Y$ consists in adding only attributes with a smaller support than any attribute from $Y$.
The most difficult case for selecting valid patterns occurs when a measure is not locally anti-monotonic.
Then, valid patterns can be retained by postfiltering, i.e., finding a (large set of) patterns satisfying an antimonotone constraint and filtering them w.r.t. the chosen nonmonotonic measure (i.e., neither monotonic nor anti-monotonic)~\cite{Roth2008,Moerchen2011,Tatti2014}, or using heuristics such as leap search~\cite{Yan2008} or low probability of finding interesting patterns in the current branch~\cite{Webb2010}.

Most of the measures are only applicable to one type of patterns, e.g., pattern leverage or cosine interest can be applied only to binary data since their definitions involve single attributes. ``Pattern independent measures'' usually relies on support of the pattern and/or on support of other patterns from the search space.
In particular, support, stability~\cite{KuznetsovStability2007}, margin-closeness~\cite{Moerchen2011} and robustness~\cite{Tatti2014} are pattern independent measures.
In this paper we work with interval tuple data, where only pattern independent measures as well as specific measures for interval tuples can be applied.
In addition, given a measure, it can be difficult to define a good threshold.
Thus various approaches for finding top-$K$ patterns were introduced~\cite{Han2002,Xin2006,Webb2011}, with the basic idea to automatically adjust the threshold for a measure $\M$.

In this paper we introduce a new algorithm \ThetaAlgoName{}, i.e., Sofia, for ``Searching for Optimal Formal Intents Algorithm'' for a interestingness threshold \texttheta, for extracting the best patterns of a kind, e.g., itemsets, interval tuples, strings, graph patterns, etc.
\ThetaAlgoName{} algorithm is applicable to a class of measures called ``projection-antimonotonic measures'' or more precisely ``measures anti-monotonic w.r.t. a chain of projections''.
This class includes globally anti-monotonic measures such as support, locally anti-monotonic measures such as cosine interest and some of the nonmonotonic measures such as stability or robustness of closed patterns.
The main novelty of this paper is \ThetaAlgoName{}, a new efficient algorithm for finding best patterns of different kinds w.r.t. projection-antimonotonic measures which constitutes a rather large class of measures.

The remaining of the paper is organized as follows.
The formalization of the current approach is based on Formal Concept Analysis (FCA)~\cite{Ganter1999} and pattern structures~\cite{Ganter2001} which are introduced in Section~\ref{sect:fca}.
Then, \ThetaAlgoName{} algorithm is detailed in Section~\ref{sect:algo} first for an arbitrary measure and second for the $\Delta$-measure.
Experiments and a discussion are proposed in Section~\ref{sect:experiment}, before conclusion.

\section{Data Model}\label{sect:fca}
\subsection{FCA and Pattern structures}\label{sect:fca-pattern-structures}

Formal Concept Analysis (FCA) is a formalism for knowledge discovery and data mining thanks to the design of concept lattices~\cite{Ganter1999}.
It is also convenient for describing models of itemset mining, and, since \cite{Pasquier1999}, lattices of closed itemsets (i.e., concept lattices) and closed descriptions are used for concise representation of association rules.
For more complex data such as sequences and graphs one can use an extension of the basic model, called  pattern structures~\cite{Ganter2001}. With pattern structures it is possible to define closed descriptions and to give a concise representation of association rules for different descriptions with a natural order (such as subgraph isomorphism order)~\cite{Kuznetsov2005,KaytoueIJCAI2011}.

	A \textit{pattern structure} is a triple $(G,(D,\sqcap),\delta)$, where $G$ is a set of objects, $(D,\sqcap)$ is a complete meet-semilattice of descriptions and $\delta: G \rightarrow D$ maps an object to a description.

\newcommand{\sqcapb}{\sqcap}

The intersection  $\sqcap$ gives the similarity of two descriptions. Standard FCA can be presented in terms of a pattern structure. A \textit{formal context} $(G,M,I)$, where $G$ is a set of objects, $M$ is a set of attributes and $I \subseteq G \times M$ an incidence relation giving information about attributes related to objects, is represented as a pattern structure $(G,(\wp(M),\cap),\delta)$, where $(\wp(M),\cap)$ is a semilattice of subsets of $M$ with $\cap$ being the set-theoretical intersection. If $x=\set{a\+b\+c}$ and $y=\set{a\+c\+d}$, then $x \sqcapb y = x \cap y = \set{a\+c}$. The mapping $\delta: G \rightarrow \wp(M)$ is given by $\delta(g)=\{m \in M \mid (g,m) \in I\}$ and returns the description of a given object as a set of attributes.

The following mappings or diamond operators give a Galois connection between the powerset of objects and descriptions:
\begin{align*}
	A^\diamond &:= \underset{g \in A}{\bigsqcap}\delta(g), &\text{for } A \subseteq G\\
	d^\diamond &:= \{g \in G \mid d \sqsubseteq \delta(g)\}, &\text{for } d \in D
\end{align*}

Given a subset of objects $A$, $A^\diamond$ returns the description which is common to all objects in $A$. Given a description $d$, $d^\diamond$ is the set of all objects whose description subsumes $d$.
A partial order $\sqsubseteq$ (subsumption) on descriptions from $D$  is defined w.r.t. the similarity operation $\sqcap$:
$c \sqsubseteq d \Leftrightarrow c \sqcap d = c$, and $c$ is subsumed by $d$.

A \textit{pattern concept} of a pattern structure $(G,(D,\sqcap),\delta)$ is a pair $(A,d)$, where $A \subseteq G$, called \textit{pattern extent} and $d \in D$, called \textit{pattern intent}, such that $A^\diamond = d$ and $d^\diamond = A$. A pattern extent is a closed set of objects, and a pattern intent is a closed description, e.g., a closed itemset when descriptions are given as sets of items (attributes).
As shown in~\cite{Kuznetsov2005}, descriptions closed in terms of counting inference (which is a standard data mining approach), such as closed graphs~\cite{Yan2003}, are elements of pattern intents.

A pattern extent corresponds to the maximal set of objects $A$ whose descriptions subsume the description $d$, where $d$ is the maximal common description for objects in $A$.
The set of all pattern concepts is partially ordered w.r.t. inclusion on extents, i.e., $(A_1,d_1)\leq (A_2,d_2)$ iff $A_1\subseteq A_2$ (or, equivalently, $d_2\sqsubseteq d_1$), making a lattice, called pattern lattice.

\subsection{Interval pattern structure}
A possible instantiation of pattern structures is interval pattern structures introduced to support efficient processing of numerical data without binarization~\cite{KaytoueIJCAI2011}. Given $k$ numerical or interval attributes whose values are of the form $[a,b]$, where $a,b \in \mathbb{R}$, the language of a pattern space is given by tuples of intervals of size $k$. For simplicity, we denote intervals of the form $[a,a]$ by $a$.

Figure~\ref{fig:stab-context} exemplifies an interval dataset. It contains 6 objects and 2 attributes. An interval as a value of an attribute corresponds to an uncertainty in the value of the  attribute. For example, the value of $m_1$ for $g_2$ is known exactly, while the value of $m_2$ is lying in $[1,2]$. Given this intuition for intervals it is natural to define similarity of two intervals as their convex hull, since by adding new objects one increases the uncertainty. For example, for $g_1$ the value of $m_1$ is 0, while for $g_6$ it is 1, thus given the set $\set{g_1,g_6}$, the uncertainty of $m_1$ in this set is $[0,1]$, i.e., the similarity of $g_1$ and $g_6$ w.r.t. $m_1$ is $[0,1]$. More formally, given two intervals $[a,b]$ and $[c,d]$, the similarity of these two intervals is given by $[a,b] \sqcap [c,d]=[\min(a,c),\max(b,d)]$. Given a tuple of intervals, the similarity is computed component-wise. For example, $g_1^\diamond \sqcap g_6^\diamond=\tuple{[0,1]}{[0,2]}$. Reciprocally, $\tuple{[0,1]}{[0,2]}=\set{g_1,g_2,\cdots,g_6}$.

The resulting concept lattice is shown in Figure~\ref{fig:stab-lattice}. Concept extents are shown by indices of objects, intents are given in angle brackets, the numbers on edges and on concepts are related to interestingness of concepts and will be described in the next subsection.

\subsection{Stability index of a concept}\label{sect:fca-stability}

\input{tex/tbl-stability-context}

For real datasets, the number of patterns can be very large, even computing the number of closed patterns is a \#P-complete problem~\cite{Kuznetsov2001}. Different measures were tested for selecting most interesting patterns,  such as  stability~\cite{KuznetsovStability2007}. Stability measures the independence of a concept intent w.r.t. randomness in data.

Given a concept $\cC$, \emph{concept stability} $\stab(\cC)$ is the relative number of subsets of the concept extent (denoted by $\extC(\cC)$), whose descriptions, i.e., the result of $(\cdot)^\diamond$ is equal to the concept intent (denoted by $\intC(\cC)$).
\begin{equation}\label{eq:stability}
	\stab(\cC):=\frac{|\{s\in\wp(\extC(\cC)) \mid s^\diamond = \intC(\cC)\}|}{|\wp(\extC(\cC))|}
\end{equation}

Here $\wp(P)$ is the powerset of $P$. The larger the stability, the more objects can be deleted from the context without affecting the intent of the concept, i.e., the intent of the most stable concepts is likely to be a characteristic pattern of a given phenomenon and not an artifact of a dataset.

We say that a concept is stable if its stability is higher than a given threshold $\theta$; a pattern $p$ is stable if there is a concept in the lattice with $p$ as the intent and the concept is stable.

\begin{example}\label{ex:stab-example}
	Figure~\ref{fig:stab-lattice} shows a lattice for the context in Figure~\ref{fig:stab-context}. Concept extents are given by their indices, i.e., $\{g_1,g_2\}$ is given by $12$. The extent of the highlighted concept $\cC$ is $\extC(\cC)=\set{g_2\+g_3\+g_4}$, thus, its powerset contains $2^3$ elements. Descriptions of 2 subsets of $\extC(\cC)$ ($\set{g_4}$ and $\emptyset$) are different from the intent of $\cC$, $\intC(\cC)=\set{m_3}$, while all other subsets of $\extC(\cC)$ have a common set of attributes equal to $\tuple{0}{[1,2]}$. So, $\stab(\cC)=\frac{2^3-2}{2^3}=0.75$. Stability of other concepts is shown in brackets.
	It should be noticed that stability of all comparable patterns for $\intC(\cC)$ in the lattice is smaller than the stability of $\cC$, which highlights the nonmonotonicity of stability.
\end{example}

Concept stability is closely related to the robustness of a closed pattern~\cite{Tatti2014}. Indeed, robustness is the probability of a closed pattern to be found in a subset of the dataset. To define this probability, the authors define a weight for every subset given as a probability of obtaining this subset by removing objects from the dataset, where every object is removed with probability $\alpha$, e.g., given a subset of objects $X \subseteq G$, the probability of the induced subset is given by $p(D_\alpha=(X,(D,\sqcap),\delta))=\alpha^{|X|}(1-\alpha)^{|G\setminus X|}$. Stability in this case is the robustness of closed pattern if the weights of subsets of the dataset are equal to $2^{-|G|}$.

The problem of computing concept stability is \#P-complete~\cite{KuznetsovStability2007}. A fast computable stability estimate was proposed in~\cite{BuzmakovICFCA2014}, where it was shown that this estimate ranks concepts almost in the same way as stability does. In particular, $ \stab(\cC) \leq 1-2^{-\Delta(\cC)}$, where $\Delta(\cC)=\underset{\cD \leq \cC}{\min}|\extC(\cC)\setminus\extC(\cD)|$,
i.e., the minimal difference in supports between concept $\cC$ and all its nearest subconcepts.
For a threshold $\theta$, patterns $p$ with $\Delta(p)\geq\theta$ are called \textDelta-stable patterns.

\begin{example}\label{ex:delta-example}
	Consider the example in Figure~\ref{fig:stab-example}. Every edge in the figure is labeled with the difference in support between the concepts this edge connects. Thus, \textDelta\,of a pattern is the minimum label of the edges going down from the concept.
	The value $\Delta((\set{g_2,g_3,g_4};\tuple{0}{[1,2]}))$ is equal to 2. Another example is $\Delta((G;\tuple{[0,1]}{[0,2]}))=2$. For this example we can also see that \textDelta-measure is not anti-monotonic either.
\end{example}

\textDelta-measure is related to the work of margin-closeness of an itemset~\cite{Moerchen2011}. In this work, given a set of patterns, e.g., frequent closed patterns, the authors rank them by the minimal distance in their support to the closest superpattern divided over the support of the pattern. In our case, the minimal distance is exactly the \textDelta-measure of the pattern.

Stability and \textDelta-measure are not anti-monotonic but rather projection-antimo\-notonic. Patterns w.r.t. such kind of measures can be mined by a specialized algorithm introduced in Section~\ref{sect:algo}. But before we should introduce projections of pattern structures in order to properly define projection-antimonotonicity and the algorithm.
\subsection{Projections of Pattern Structures}

 The approach proposed in this paper is based on projections introduced for reducing complexity of computing pattern lattices~\cite{Ganter2001}.

	A \textit{projection} $\psi: D \rightarrow D$ is an ``interior operator'', i.e., it is (1)~monotonic ($x \sqsubseteq y \Rightarrow \psi(x) \sqsubseteq \psi(y)$), (2)~contractive ($\psi(x) \sqsubseteq x$) and (3)~idempotent ($\psi(\psi(x))=\psi(x)$).
 A \emph{projected pattern structure} $\psi((G,(D,\sqcap),\delta))$ is a pattern structure $(G,(D_\psi,\sqcap_\psi),\psi \circ \delta)$, where $D_\psi=\psi(D)=\{d\in D \mid \exists d^* \in D: \psi(d^*) = d\}$ and $\forall x,y \in D, x \sqcap_\psi y := \psi( x \sqcap y )$.

\begin{example}
	Consider the example in Figure~\ref{fig:stab-example}. If we remove a column corresponding to an attribute, e.g., the attribute $m_2$, from the context in Figure~\ref{fig:stab-context}, we define a projection, given by $\psi(\tuple{[a,b]}{[c,d]})=\tuple{[a,b]}{[-\infty,+\infty]}$, meaning that no value of $m_2$ is taken into account.
\end{example}

Given a projection $\psi$ we call $\psi(D)=\{d \in D \mid \psi(d)=d\}$ the \emph{fixed set of} $\psi$. Note that, if $\psi(d) \neq d$, then there is no other $\tilde{d}$ such that $\psi(\tilde{d}) = d$ because of idempotency of projections. Hence, any element outside the fixed set of the projection $\psi$ is pruned from the description space. Given the notion of a fixed set we can define a partial order on projections.

\begin{definition}\label{def:projection-order}
	Given a pattern structure $\PS=(G,(D,\sqcap),\delta)$ and two projections $\psi_1$ and $\psi_2$, we say that $\psi_1$ is simpler than $\psi_2$ ($\psi_2$ is more detailed than $\psi_1$), denoted by $\psi_1 < \psi_2$, if $\psi_1(D) \subset \psi_2(D)$, i.e., $\psi_1$ prunes more descriptions than $\psi_2$.
\end{definition}

Our algorithm is based on this order on projections. The simpler a projection $\psi$ is, the less patterns we can find in $\psi(\PS)$, and the less computational efforts one should take. Thus, we compute a set of patterns for a simpler projection, then we remove unpromising patterns and extend our pattern structure and the found patterns to a more detailed projection. This allows us to reduce the size of patterns within a simpler projection in order to reduce the computational complexity of more detailed projection.

\subsection{Projections of Interval Pattern Structures}
Let us first consider interval pattern structures with only one attribute $m$. Let us denote by $W=\set{w_1,\cdots,w_{|W|}}$ all possible values of the left and right endpoints of the intervals corresponding to the attribute in a dataset, so that $w_1 < w_2 <\cdots<w_{|W|}$. By reducing the set $W$ of possible values for the left or the right end of the interval we define a projection. For example, if $\{w_1\}$ is the only possible value for the left endpoint of an interval and $\{w_{|W|}\}$ is the only possible value of the right endpoint of an interval, then all interval patterns are projected to $[w_1,w_{|W|}]$. Let us consider this in more detail.

Let two sets $L,R \subset W$ such that $w_1 \in L$ and $w_{|W|} \in R$ be constraints on possible values on the left and right endpoints of an interval, respectively. Then a projection is defined as follows:

\begin{equation} \label{eq:interval-projection}
	\psi_{m[L,R]}([a,b])=\left[{\max}{\set{l \in L | l \leq a}}, {\min}{\set{r \in R | r \geq b}}\right].
\end{equation}

Requiring that $w_1 \in L$ and $w_{|W|} \in R$ we ensure that the sets used for minimal and maximal functions are not empty.
It is not hard to see that (\ref{eq:interval-projection}) is a projection. The projections given by (\ref{eq:interval-projection}) are ordered w.r.t. simplicity (Definition~\ref{def:projection-order}). Indeed, given $L_1\subseteq L$ and $R_1 \subseteq R$, we have $\psi_{m[L_1,R_1]} < \psi_{m[L,R]}$, because of inclusion of fixed sets.
Let us notice that a projection $\psi_{m[W,W]}$ does not modify the lattice of concepts for the current dataset, since any interval for the value set $W$ is possible.
We also notice that a projection $\psi_{m[L,R]}$ is defined for one interval, while we can combine the projections for different attributes in a tuple to a single projection for the whole tuple $\psi_{m_1[L_1,R_1]m_2[L_2,R_2]\dots}$.

\begin{example}\label{ex:ips-projection}
	Consider example in Figure~\ref{fig:stab-example}. Let us consider a projection \[
		\psi_{m_1[\{0,1\},\{1\}]m_2[\{0,2\},\{0,2\}]}.
	\]The fixed set of this projection consists of $\{[0,1],1\}\times\{0,2,[0,2]\}$, i.e., 6 intervals. Let us find the projection of $(g_2)^\diamond=\tuple{0}{[1,2]}$ in a component-wise way: $\psi_{m_1[\{0,1\},\{1\}]}(0)=[0,1]$, since 0 is allowed on the left endpoint of an interval but not allowed to be on the right endpoint of an interval; $\psi_{m_2[\{0,2\},\{0,2\}]}([1,2])=[0,2]$ since 1 is not allowed on the left endpoint of an interval. Thus, \[
		\psi_{m_1[\{0,1\},\{1\}]m_2[\{0,2\},\{0,2\}]}(\tuple{0}{[1,2]}) = \tuple{[0,1]}{[0,2]}.
	\]The lattice corresponding to this projection is shown in Figure~\ref{fig:proj-lattice}.
\end{example}

\input{tex/fig-proj-lattice}

\section{\ThetaAlgoName{} Algorithm}\label{sect:algo}
\subsection{Anti-monotonicity w.r.t. a Projection}

Our algorithm is based on the projection-antimonotonicity, a new idea introduced in this paper. Many interestingness measures for patterns, e.g., stability, are not (anti-)monotonic w.r.t. subsumption order on patterns. A measure $\M$ is called \emph{anti-monotonic}, if for two patterns $q \sqsubseteq p$, $\M(q) \geq \M(p)$. For instance, support is a anti-monotonic measure w.r.t. pattern order and it allows for efficient generation of patterns with support larger than a threshold~\cite{Agrawal1994,Mannila1994,Pasquier1999}. The projection-antimonotonicity is a generalization of standard anti-monotonicity and allows for efficient work with a larger set of interestingness measures.

\begin{definition}\label{def:monotonic-measure}
	Given a pattern structure $\PS$ and a projection $\psi$, a measure $\mathcal{M}$ is called \emph{anti-monotonic w.r.t. the projection} $\psi$, if
	\begin{equation}\label{eq:monotonic-measure}
		(\forall p \in \psi(\PS))(\forall q \in \PS, \psi(q)=p)~\mathcal{M}_\psi(p) \geq \mathcal{M}(q),
	\end{equation}
	where $\M_\psi(p)$ is the measure $\M$ of a pattern $p$ computed in $\psi(\PS)$.
\end{definition}

Here, for any pattern $p$ of a projected pattern structure we check that a preimage $q$ of $p$ for $\psi$ has a measure smaller than the measure of $p$. It should be noticed that a measure $\M$ for a pattern $p$ can yield different values if $\M$ is computed in $\PS$ or in $\psi(\PS)$. Thus we use the notation $\M_\psi$ for the measure $\M$ computed in $\psi(\PS)$.
The property of a measure given in Definition~\ref{def:monotonic-measure} is called projection-antimonotonicity.

It should be noticed that classical anti-monotonic measures are projection-antimonotonic for any projection. Indeed, because of contractivity of $\psi$ ($\psi(p) \sqsubseteq p$), for any anti-monotonic measure one has $\M(\psi(p)) \geq \M(p)$. This definition covers also the cases where a measure $\M$ is only locally anti-monotonic, i.e., given a pattern $p$ there is an immediate subpattern $q \prec p$ such that $\M(q) \geq \M(p)$, see e.g., the cosine interest of an itemset, which is only locally anti-monotonic~\cite{Cao2014}. Moreover, this definition covers also some measures that are not locally anti-monotonic. As we mentioned in Examples~\ref{ex:stab-example}~and~\ref{ex:delta-example} stability and \textDelta-measure are not locally anti-monotonic. However, it can be shown that they are anti-monotonic w.r.t. any projection~\cite{Buzmakov2013a}.
Moreover, following the same strategy one can prove that robustness of closed patterns from~\cite{Tatti2014} is also anti-monotonic w.r.t. any projection. In particular, the robustness of closed patterns defines a anti-monotonic constraint w.r.t. any projection.


Thus, given a measure $\M$ anti-monotonic w.r.t. a projection $\psi$, if $p$ is a pattern such that $\M_\psi(p) < \theta$, then $\M(q)<\theta$ for any preimage $q$ of $p$ for $\psi$. Hence, if, given a pattern $p$ of $\psi(\PS)$, one can find all patterns $q$ of $\PS$ such that $\psi(q)=p$, it is possible to first find all patterns of $\psi(\PS)$ and then to filter them w.r.t. $\M_\psi$ and a threshold, and finally to compute the preimages of filtered patterns. It allows one to cut earlier unpromising branches of the search space or adjust a threshold for finding only a limited number of best patterns.
\subsection{Anti-monotonicity w.r.t. a Chain of Projections}

However, given just one projection, it can be hard to efficiently discover the patterns, because the projection is either hard to compute or the number of unpromising patterns that can be pruned is not high. Hence we introduce \emph{a chain of projections} $\psi_0 < \psi_1 < \cdots < \psi_k=\mathbb{1}$, where a pattern lattice for $\psi_0(\PS)$ can be easily computed and $\mathbb{1}$ is the identity projection, i.e., $(\forall x)\mathbb{1}(x)=x$. For example, to find frequent itemsets, we typically search for small frequent itemsets and then extend them to larger ones. This corresponds to extension to a more detailed projection.


\begin{definition}\label{def:monotonic-projection-chain}
	Given a pattern structure $\PS$ and a chain of projections $\psi_0 < \psi_1 < \cdots < \psi_k=\mathbb{1}$, a measure $\M$ is called \emph{anti-monotonic w.r.t. the chain of projections} if $\M$ is anti-monotonic w.r.t. all $\psi_i$ for $0 \leq i \leq k$.
\end{definition}

\begin{example}\label{ex:ips-projection-chain}
	Let us construct a chain of projections satisfying~(\ref{eq:interval-projection}) for the example in Figure~\ref{fig:stab-example}. The value set for the first attribute is $W_1=\set{0,1}$ and the value set for the second is $W_2=\set{0,1,2}$. Let us start the chain from a projection $\psi_0=\psi_{m_1[\{0\},\{1\}]m_2[\{0\},\{2\}]}$. This projection allows only for one pattern $\tuple{[0,1]}{[0,2]}$, i.e., the concept lattice is easily found. Then we increase the complexity of a projection by allowing more patterns. For example, we can enrich the first component of a tuple without affecting the second one, i.e., a projection $\psi_1=\psi_{m_1[\{0,1\},\{0,1\}]m_2[\{0\},\{2\}]}$. This projection allows for 3 patterns, i.e., any possible interval of the first component and only one interval [0,2] for the second component. Let us notice that it is not hard to find preimages for $\psi_0$ in $\psi_1(D)$. Indeed, for any pattern $p$ from $\psi_0(D)$ one should just modify either the left side of the first interval of $p$ by one value, or the right side of the first interval of $p$.

Then we can introduce a projection that slightly enrich the second component of a tuple, e.g., $\psi_2=\psi_{m_1[\{0,1\},\{0,1\}]m_2[\{0,1\},\{1,2\}]}$ and finally we have $\psi_3=\psi_{m_1[W_1,W_1]m_2[W_2,W_2]}$. Finding preimages in this chain is not a hard problem, since on every set we can only slightly change left and/or right side of the second interval in a tuple. Thus, starting from a simple projection and making transitions from one projection to another, we can cut unpromising branches and efficiently find the set of interesting patterns.
\end{example}

\subsection{Algorithms}

\begin{algorithm}[th]
	\AlgoDisplayBlockMarkers
	\SetAlgoBlockMarkers{}{}
	\SetAlgoNoEnd
	\SetKwProg{Fn}{Function}{}{}
	\SetKwFunction{ExtProj}{ExtendProjection}
	\SetKwFunction{Preimages}{Preimages}
	\SetKwFunction{ThetaAlgo}{Algorithm\_\ThetaAlgoName{}}
	\SetKwFunction{FindPatterns}{FindPatterns}
	
	\KwData{
		A pattern structure $\PS$, a chain of projections $\Psi=\set{\psi_0,\psi_1,\cdots,\psi_k}$, a measure $\M$ anti-monotonic for the chain $\Psi$, and a threshold $\theta$ for $\M$.
	}
	\Fn{\ExtProj{$i$, $\theta$, $\mathcal{P}_{i-1}$}}{
		\KwData{
			$i$ is the projection number to which we should extend ($0<i\leq k$), $\theta$ is a threshold value for $\M$, and $\mathcal{P}_{i-1}$ is the set of patterns for the projection $\psi_{i-1}$.
		}
		\KwResult{
			The set $\mathcal{P}_i$ of all patterns with the value of measure $\M$ higher than the threshold $\theta$ for $\psi_i$.
		}
		$\mathcal{P}_{i}\longleftarrow\emptyset$\;
		\tcc{Put all preimages in $\psi_i(\PS)$ for any pattern $p$}
		\ForEach{$p \in \mathcal{P}_{i-1}$}{
			$\mathcal{P}_i\longleftarrow\mathcal{P}_i \cup \Preimages{i,p}$	
		}
		\tcc{Filter patterns in $\mathcal{P}_i$ to have a value of $\M$ higher than $\theta$}
		\ForEach{$p \in \mathcal{P}_i$}{
			\If{$\M_{\psi_i}(p) \leq \theta$}{
				$\mathcal{P}_i \longleftarrow \mathcal{P}_i \setminus \set{p}$
			}
		}
	}
	\Fn{\ThetaAlgo}{
		\KwResult{
			The set $\mathcal{P}$ of all patterns with a value of $\M$ higher than the threshold $\theta$ for $\PS$.
		}
		\tcc{Find all patterns in $\psi_0(\PS)$ with a value of $\M$ higher than $\theta$}
		$\mathcal{P}\longleftarrow\FindPatterns{$\theta,\psi_0$}$\;
		\tcc{Run through out the chain $\Psi$ and find the patterns for $\psi_i(\PS)$}
		\ForEach{$0 < i \leq k$}{
			$\mathcal{P}\longleftarrow\ExtProj{$i, \theta, \mathcal{P}$}$\;
		}
	}
	\caption{The \ThetaAlgoName{} algorithm for finding patterns in $\PS$ with a value of a measure $\M$ higher than a threshold $\theta$.  \label{alg:theta-sofia}}
\end{algorithm}

Given a measure anti-monotonic w.r.t. a chain of projections, if we are able to find all preimages of any element in the fixed set of $\psi_i$ that belong to a fixed set of $\psi_{i+1}$, then we can find all patterns of $\PS$ with a value of $\M$ higher than a given threshold $\theta$. We call this algorithm \ThetaAlgoName{} and its pseudocode is given in Algorithm~\ref{alg:theta-sofia}. In lines 11-12 we find all patterns for $\psi_0(\PS)$ satisfying the constraint that a value of $\M$ is higher than a threshold. Then in lines 13-15 we iteratively extend projections from simpler to more detailed ones. The extension is done by constructing the set $\mathcal{P}_i$ of preimages of the set $\mathcal{P}_{i-1}$ (lines 2-5) and then by removing the patterns that do not satisfy the constraint from $\mathcal{P}_i$ (lines 6-9).

The algorithm is sound and complete, since first, a pattern $p$ is included into the set of preimages of $p$ ($\psi(p)=p$) and second, if we remove a pattern $p$ from the set $\mathcal{P}$, then the value $\M(p) < \theta$ and, hence, the measure value of any preimage of $p$ is less than $\theta$ by the projection-antimonotonicity of $\M$.
The worst case time complexity of \ThetaAlgoName{} algorithm is
{\small
\begin{align}\label{eq:complexity-theta}
	\mathbb{T}(\text{\ThetaAlgoName})&=\mathbb{T}(FindPatterns(\psi_0))+\notag\\
	&\hspace{-1mm}+k\cdot\underset{0<i\leq k}{\max}|\mathcal{P}_i|\cdot(\mathbb{T}(Preimages)+\mathbb{T}(\M)),
\end{align}
}where $k$ is the number of projections in the chain, $\mathbb{T}(\mathcal{X})$ is time for computing operation $\mathcal{X}$.
Since projection $\psi_0$ can be chosen to be very simple, in a typical case the complexity of $FindPatterns(\theta,\psi_0)$ can be low or even constant. The complexities of $Preimages$ and $\M$ depend on the measure, the chain of projections, and the kind of patterns.
In many cases $\underset{0<i\leq k}{\max}|\mathcal{P}_i|$ can be exponential in the size of the input, because the number of patterns can be exponential.
It can be a difficult task to define the threshold $\theta$ such that the maximal cardinality of $\mathcal{P}_i$ is not larger than a given number.
This can be solved by an automatically adjustment of the threshold $\theta$, which is not discussed here.
\subsection{\ThetaAlgoName{} Algorithm for Interval Tuple Data}
In this subsection we consider a pattern structure $\K=(G,(D_I,\sqcap),\delta)$, where $D_I$ is a semilattice of interval tuple descriptions. We say that every component of a tuple $p$ corresponds to an attribute $m \in M$, where $M$ is the set of interval attributes. Thus, the size of any tuple in $D_I$ is $|M|$, and for any attribute $m \in M$ we can denote the corresponding interval by $m(p)$.
We also denote the value set of $m$ by $W_m$.
Since the set $W_m$ is totally ordered we also denote by $W_m^{(j)}$ and $W_m^{(-j)}$ the sets containing the first $j$ (smallest) elements and the last j (largest) elements from $W_m$, respectively.

A projection chain for interval tuple data is formed in the same way as discussed in Example~\ref{ex:ips-projection-chain}. We start from the projection containing only one pattern corresponding to the largest interval in each component, i.e., for an attribute $m$ the projection is of the form $\psi_m[W_m^{(1)},W_m^{(-1)}]$. Then to pass to a next projection, we select the attribute $m$, and for this attribute we extend the projection from $\psi_m[W_m^{(j)},W_m^{(-j)}]$ to $\psi_m[W_m^{(j+1)},W_m^{(-j-1)}]$. Thus, there are $k=\underset{m \in M}{\max}|W_m|\cdot |M|$ projections.



Finding preimages in this case is not hard, since to make a projection more detailed one should just extend the corresponding interval in left and/or on right end of the interval, i.e., there are only 4 possible preimages for a pattern when passing from one projection to another in this chain. Thus, we have proved the following

\begin{proposition}
The worst case complexity for \ThetaAlgoName{} algorithm for interval tuple data is
{\small
\begin{align}\label{eq:complexity-theta-IPS}
	\mathbb{T}(\text{\ThetaAlgoName}_{\text{intervals}})=\underset{m \in M}{\max}|W_m|\cdot |M| \cdot\underset{0<i\leq k}{\max}|\mathcal{P}_i|\cdot\mathbb{T}(\M).
\end{align}
}.
\end{proposition}

%
\subsection{\ThetaAlgoName{} Algorithm for Closed Patterns}\label{sect:closed-patterns}
Closed frequent itemsets are widely used as a condensed representation of all frequent itemsets since~\cite{Pasquier1999}.  Here we show how we can adapt the algorithm for closed patterns.
A closed pattern in $\psi_{i-1}(\PS)$ is not necessarily closed in $\psi_i(\PS)$. However, the extents of $\psi(\PS)$ are extents of $\PS$~\cite{Ganter2001}. Thus, we associate the closed patterns with extents and then work with extents instead of patterns, i.e., a pattern structure $\PS=(G,(D,\sqcap),\delta)$ is transformed into $\PS_C=(G,(D_C,\sqcap_C),\delta_C)$, where $D_C=2^G$. Moreover, for all $x,y \in D_C$ we have $x \sqcap_C y = (x^\diamond \sqcap y^\diamond)^\diamond$, where diamond operator is computed in $\PS$ and $\delta_C(g \in G)=\set{g}$. Hence, every pattern $p$ in $D_C$ corresponds to a closed pattern $p^\diamond$ in $D$.
A projection $\psi$ of $\PS$ induces a projection $\psi_C$ of $\PS_C$, given by $\psi_C(X \subseteq G)=\psi(X^\diamond)^\diamond$ with $(\cdot)^\diamond$ for $\PS$.

%
\subsection{\textDelta-measure and \ThetaAlgoName{} Algorithm}\label{sect:measures}

In this subsection we show that \textDelta-measure is anti-monotonic for any projection; it is a stronger condition than the one required by Definition~\ref{def:monotonic-projection-chain}. \textDelta-measure works for closed patterns, and, hence, we identify every description by its extent (Subsection~\ref{sect:closed-patterns}).

\begin{proposition}
$\Delta$ is anti-monotonic for any projection $\psi$.
\end{proposition}
\emph{Proof.} By properties of a projection,  an extent of $\psi(\PS)$ is an extent of $\PS$~\cite{Ganter2001}. Let us consider an extent $E$ and an extent of its descendant in $\psi(\PS)$. Let us suppose that $E_p$ is a preimage of $E$ for the projection $\psi$. Since $E_c$ and $E_p$ are extents in $\PS$, the set $E_{cp}=E_c \cap E_p$ is an extent in $\PS$ (the intersection of two closed sets is a closed set). Since $E_p$ is a preimage of $E$, then $E_p \not\leq E_c$ (otherwise, $E_p$ is a preimage of $E_c$ and not of $E$). Then, $E_{cp} \neq E_p$ and $E_{cp} \leq E_p$. Hence, $\Delta(E_{p})\leq |E_{p} \setminus E_{cp}| \leq |E \setminus E_c|$. So, given a preimage $E_p$ of $E$, $(\forall E_c \subseteq E)\Delta(E_{p}) \leq |E \setminus E_c|$, i.e., $\Delta(E_p) \leq \Delta(E)$. Thus, we can use \textDelta-measure in combination with \ThetaAlgoName{}.

\subsection{Example of \textDelta-Stable Patterns in Interval Tuple Data}\label{sect:example}

\input{tex/tbl-ex-algorithm-concepts}

Let us consider the example in Figure~\ref{fig:stab-example} and show how we can find all \textDelta-stable patterns with a threshold $\theta=2$. The chain of projections for this example is given in Example~\ref{ex:ips-projection-chain}, it contains 4 projections:
\begin{align*}
	&\psi_0=\psi_{m_1[\{0\},\{1\}]m_2[\{0\},\{2\}]}
	& \psi_1=\psi_{m_1[\{0,1\},\{0,1\}]m_2[\{0\},\{2\}]}\\
	&\psi_2=\psi_{m_1[\{0,1\},\{0,1\}]m_2[\{0,1\},\{1,2\}]}
	& \psi_3=\psi_{m_1[\{0,1\},\{0,1\}]m_2[\{0,1,2\},\{0,1,2\}]}
\end{align*}

Since we are looking for closed patterns, every pattern can be identified by its extent. In Table~\ref{tab:ex-algorithm-concepts} all patterns are given by their extents, i.e., by elements of $D_C$. For every pattern \textDelta-measure is shown for every $\psi_i$. A cell is shown in grey if the pattern is no more considered (the value of \textDelta{} less than 2). A cell has a dash ``--'', if a pattern in the row has not been generated for this projection.

For the example in Figure~\ref{fig:stab-example} the global process is as follows. At the beginning $\psi_0(D_I)$ contains only one element corresponding to pattern extent $123456$ (a short cut for $\{g_1,g_2,g_3,g_4,g_5,g_6\}$) with a description $\tuple{[0,1]}{[0,2]}$. Then, in $\psi_1(G,(D_I,\sqcap),\delta)$ possible preimages of $123456$ are patterns with descriptions $\tuple{0}{[0,2]}$ and $\tuple{1}{[0,2]}$ given by pattern extents $1234$ and $56$, respectively. Then we continue with these three patterns which are all \textDelta-stable for the moment. The pattern extents $123456$ and $56$ have no preimages for the transition $\psi_1 \rightarrow \psi_2$, while the pattern extent $1234$ has two preimages with descriptions $\tuple{0}{[0,1]}$ and $\tuple{0}{[1,2]}$ for this projection, which correspond to pattern extents $1$ and $234$. The first one is not \textDelta-stable and thus is no more considered. Moreover, the pattern extent $1234$ is not \textDelta-stable (because of $234$) and should also be removed. Finally, in transition $\psi_2 \rightarrow \psi_3$ only extent-pattern $234$ has a preimage, a pattern extent $4$, which is not \textDelta-stable. In such a way, we have started from a very simple projection $\psi_0$ and achieved the projection $\psi_3$ that gives us the \textDelta-stable patterns of the target pattern structure.

\section{Experiments and Discussion}\label{sect:experiment}

In this section we compare our approach to approaches based on postfiltering. Indeed, there is no approach that can directly mine stable-like pattern, e.g., stable, \textDelta-stable or robust patterns. The known approaches use postfiltering to mine such kind of patterns~\cite{Roth2008,Moerchen2011,Buzmakov2013a,Tatti2014}. Recently it was also shown that it is more efficient to mine interval tuple data without binarization~\cite{KaytoueIJCAI2011}. In their paper the authors introduce algorithm \texttt{MinIntChange} for working directly with interval tuple data. Thus we compare \ThetaAlgoName{} and \texttt{MinIntChange} for finding \textDelta-stable patterns. We find \textDelta-stable concepts with \ThetaAlgoName{} and then adjust frequency  threshold $\theta$ such that all \textDelta-stable patterns are among the frequent ones.

The experiments are carried out on an ``Intel(R) Core(TM) i7-2600 CPU @ 3.40GHz'' computer with 8Gb of memory under Ubuntu 14.04 operating system. The algorithms are not parallelized and are coded in \texttt{C++}.

\subsection{Dataset Simplification}

For interval tuple data stable patterns can be very deep in the search space, such that neither of the algorithms can find them quickly. Thus, we join some similar values for every attribute in an interval in the following way. Given a threshold $0 < \beta$, two consequent numbers $w_i$ and $w_{i+1}$  from a value set $W$ are joined in the same interval if $w_{i+1} - w_i < \beta$. In order to properly set the threshold $\beta$, we use another threshold $0 < \gamma < 1$, which is much easier to set.

If we assume that the values of the attribute $m$ are distributed around several states with centers $\tilde{w}^1,\cdots,\tilde{w}^l$, then it is natural to think that the difference between the closest centers $\mathtt{abs}(\tilde{w}^i - \tilde{w}^{i \pm 1})$ are much larger than the difference between the closest values. Ordering all values in the increasing order and finding the maximal difference $\delta_{\max}$ can give us an idea of typical distance between the states in the data. Thus, $\gamma$ is defined as a proportion of this distance that should be considered as a distance between states, i.e., we put $\beta = \gamma \cdot \delta_{\max}$. If the distance between closest values in $W$ are always the same, then even $\gamma =0.99$ does not join values in intervals. However, if there are two states and the values are distributed very closely to one of these two states, then even $\gamma=0.01$ can join values into one of two intervals corresponding to the states.

\subsection{Datasets}

We take several datasets from the Bilkent University database~\footnote{\url{http://funapp.cs.bilkent.edu.tr/DataSets/}}. The datasets are summarized in Table~\ref{tbl:experiments}. The names of datasets are given by standard abbreviations used in the database of Bilkent University. For every dataset we provide the number of objects and attributes and the threshold $\gamma$ for which the experiments are carried out. For example, database \texttt{EM} has 61 objects, 9 numeric attributes, and the threshold $\gamma$ is set to 0.3.
Categorical attributes and rows with missing values, if any, are removed from the datasets.

\subsection{Experiments}

In Table~\ref{tbl:experiments} we show the computation time for finding the best \textDelta-stable pattern (or patterns if they have the same value for \textDelta-measure) for \ThetaAlgoName{} and for \texttt{MinIntChange}. The last algorithm is abbreviated as \texttt{MIC}. Since \texttt{MinIntChange} algorithm sometimes produces too many patterns, i.e., we do not have enough memory in our computer to check all of them, we interrupt the procedure and show the corresponding time in grey. We also show the number of the best patterns and the corresponding threshold $\Delta$. The support threshold $\theta$ for finding the best \textDelta-stable patterns is also shown.
For example, dataset \texttt{CN} contains 5362 best \textDelta-stable patterns, all having a \textDelta\, of 2.
To find all these patterns with a postfiltering, we should mine frequent patterns with a support threshold lower than 30 or $\frac{30}{105}=30\%$.
\ThetaAlgoName{} computes all these patterns in 2.4 seconds, while \MIC requires at least 28 seconds and the procedure was interrupted without continuation.

As we can see, \ThetaAlgoName{} is significantly faster than \MIC in all datasets.
In the two datasets \texttt{CA} and \texttt{PT}, \MIC was stopped before computing all patterns and the runtime did not exceed the runtime of \ThetaAlgoName{}.
However, in both cases, \MIC achieved less than 10\% of the required operations.

\input{tex/tbl-experiments}

\section{Conclusion}

In this paper we have introduced a new class of interestingness measures that are anti-monotonic w.r.t. a chain of projections. We have designed a new algorithm, called \ThetaAlgoName{}, which is able to efficiently find the best patterns w.r.t. such interestingness measures for interval tuple data. The experiments reported in the paper are the witness of the efficiency of the \ThetaAlgoName{} algorithms compared to indirect approaches based on postfiltering.
Many future research directions are possible.  Different measures should be studied in combination with \ThetaAlgoName{}. One of them is robustness, which is very close to stability and can be applied to nonbinary data.
Moreover, the choice of a projection chain is not a simple one and can affect the algorithm efficiency.
Thus, a deep study of suitable projection chains should be carried out.

\vspace{3mm}
{\noindent\small
\textbf{Acknowledgments:}
this research was supported by the Basic Research Program at the National Research University Higher School of Economics (Moscow, Russia) and by the BioIntelligence project (France).
\vspace{-2mm}
}

\bibliographystyle{plain}
\putBibliography

\end{document}

%% file: tex/packages.tex
\usepackage[utf8x]{inputenc}
\usepackage{ucs}

\usepackage{amsmath}
\usepackage{amsfonts}
\usepackage{amssymb}
\usepackage{stmaryrd} 
\usepackage{bbold} 

\IfFileExists{subcaption.sty}{
	\usepackage{subcaption}
	\captionsetup{compatibility=false}
}{
	\usepackage{caption}
	\DeclareCaptionSubType{figure}
	\makeatletter
	\caption@For{subtypelist}{%
		\newenvironment{sub#1}%
		{\caption@withoptargs\subcaption@minipage}%
		{\endminipage}}%
	\newcommand*\subcaption@minipage[2]{%
		\minipage#1{#2}%
		\setcaptionsubtype\relax}
	\makeatother
}

\usepackage{url}
\usepackage{afterpage}

\usepackage{multirow}
\usepackage{colortbl}
\usepackage{tablefootnote}

\usepackage{tikz}
\usetikzlibrary{calc,positioning}

\usepackage[lined,boxed,commentsnumbered,linesnumbered]{algorithm2e}

\usepackage{textgreek}
\usepackage[lined,boxed,commentsnumbered,linesnumbered]{algorithm2e}

\usepackage{xspace}

%% file: tex/settings.tex
\newcommand{\putBibliography}{
	\bibliography{library.bib,/home/leshyk/PhD/library.bib}
}
\SetAlCapSkip{10pt}

%% file: tex/commands.tex
\newcommand{\PS}{\mathbb{P}} 
\newcommand{\K}{\mathbb{K}}

\newcommand{\cC}{\mathcal{C}}
\newcommand{\cD}{\mathcal{D}}
\newcommand{\stab}{\mathtt{Stab}}
\newcommand{\extC}{\mathtt{Ext}}
\newcommand{\intC}{\mathtt{Int}}

\newcommand{\set}[1]{\{#1\}}

\newcommand{\M}{\mathcal{M}}

\newcommand{\AlgoName}{\textSigma\textomikron\textphi\textiota\textalpha}
\newcommand{\ThetaAlgoName}{\texttheta-\textSigma\textomikron\textphi\textiota\textalpha}

\newcommand{\MIC}{\texttt{MinIntChange}\xspace}

%% file: tex/graphicks.tikz.tex
\newcommand{\tikznamedpicture}[3][0]{
	\newcommand{#2}[#1]{ \begin{tikzpicture}#3\end{tikzpicture} }
}

\newcommand{\concept}[2]{$\left(#1;#2\right)$}
\newcommand{\tuple}[2]{\left<#1;#2\right>}

\tikznamedpicture{\putStabLattice}{[
		every node/.style={draw,rectangle,font={\scriptsize}},
		node distance= 0.6cm and 0.3cm
	]
	\node(bot) {\concept{\emptyset}{\top}\bf[1]};
	\node(g4)[above=of bot] {\concept{4}{\tuple{0}{2}}\bf[0.5]}
		edge node [left,draw=none] {1} (bot);
	\node(g1)[left=of g4] {\concept{1}{\tuple{0}{0}}\bf[0.5]}
		edge node [left,draw=none] {1} (bot);
	\node(g234)[above=of g4,very thick] {\concept{234}{\tuple{0}{[1,2]}}\bf[0.75]}
		edge node [left,draw=none] {2} (g4);
	\node(g1234)[above=of $(g234)!0.5!(g234-|g1)$] {\concept{1234}{\tuple{0}{[0,2]}}\bf[0.44]}
		edge node [right,draw=none] {1} (g234)
		edge node [left,draw=none] {3} (g1);
	\node(g56)[right=of g234]{\concept{56}{\tuple{1}{[0,2]}}\bf[0.75]}
		edge node [right,draw=none] {2} (bot);
	\node(g123456)[above=of $(g1234)!0.5!(g1234-|g56)$]{\concept{123456}{\tuple{[0,1]}{[0,2]}}\bf[0.7]}
		edge node [xshift=-1mm,left,draw=none] {2} (g1234)
		edge node [right,draw=none] {4} (g56);
}
\tikznamedpicture{\putProjLattice}{[
		every node/.style={draw,rectangle,font={\scriptsize}},
		node distance= 0.6cm and 0.3cm
	]
	\node(bot) {\concept{\emptyset}{\top}};
	\node(g4)[above=of bot] {\concept{4}{\tuple{[0,1]}{2}}}
		edge  (bot);
	\node(g1)[left=of g4] {\concept{1}{\tuple{[0,1]}{0}}}
		edge  (bot);
	\node(g56)[right=of g4]{\concept{56}{\tuple{1}{[0,2]}}}
		edge  (bot);
	\node(g123456)[above=of g4]{\concept{123456}{\tuple{[0,1]}{[0,2]}}}
		edge  (g4)
		edge  (g1)
		edge  (g56);
}

%% file: tex/tbl-stability-context.tex
\begin{figure}
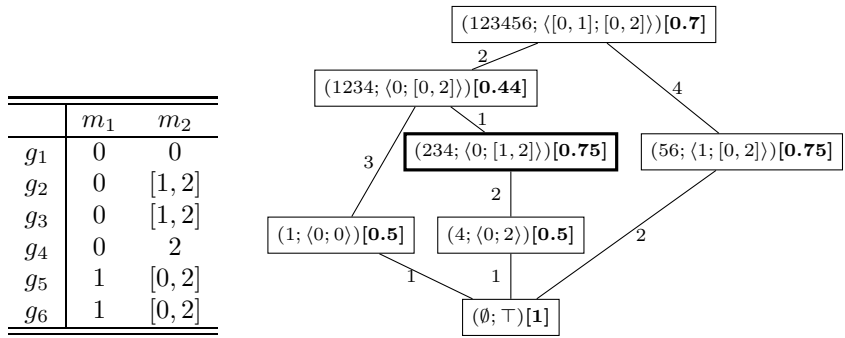

	\centering
	\begin{subfigure}[b]{0.3\columnwidth}
		\centering
		\begin{tabular}{l|cc}
			\hline\hline
			& $m_1$ & $m_2$ \\
			\hline
			$g_1$
			& $0$ & $0$ \\
			$g_2$ 
			& $0$ & $[1,2]$ \\
			$g_3$ 
			& $0$ & $[1,2]$ \\
			$g_4$ 
			& $0$ & $2$ \\
			$g_5$ 
			& $1$ & $[0,2]$ \\
			$g_6$ 
			& $1$ & $[0,2]$ \\
			\hline\hline 
		\end{tabular}
		\caption{An interval context.}
		\label{fig:stab-context}
		\vspace{4mm}
	\end{subfigure}
	\begin{subfigure}[b]{0.65\columnwidth}
		\centering
		\putStabLattice
		\caption{An interval concept lattice with corresponding stability indexes. Objects are given by their indices.}
		\label{fig:stab-lattice}
	\end{subfigure}
	\caption{A formal context and the corresponding lattice.}
	\label{fig:stab-example}
\end{figure}

%% file: tex/fig-proj-lattice.tex
\begin{figure}[th]
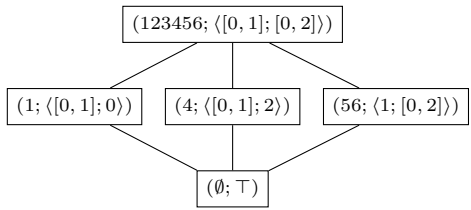

	\centering
	\putProjLattice
	\caption{Projected lattice from example in Figure~\ref{fig:stab-example} by projection $\psi_{m_1[\{0,1\},\{1\}]m_2[\{0,2\},\{0,2\}]}$. See Example~\ref{ex:ips-projection}.}
	\label{fig:proj-lattice}
\end{figure}

%% file: tex/tbl-ex-algorithm-concepts.tex
\newcommand{\g}{\cellcolor[gray]{.7}}
\begin{table}[t]
	\centering
	\caption{Patterns found for every projection in a chain for the example in Figure~\ref{fig:stab-example}. Patterns are grey if they are removed for the corresponding projetion and they are labeled with ``--'' if they have not yet been found.}
	{
	\begin{tabular}{l|c|cccc}
		\hline\hline
		\multirow{2}{*}{\#} & \multirow{2}{*}{Pattern Ext.} & \multicolumn{4}{c}{\textDelta-measure} \\
		& & $\psi_0$ & $\psi_1$ & $\psi_2$ & $\psi_3$ \\
		\hline
		1 & $\set{g_1,g_2,g_3,g_4,g_5,g_6}$
		&  6  &   2  &   2  &   2  \\
		2 & $\set{g_1,g_2,g_3,g_4}$
		& --  &   4  &\g 1  &\g 1  \\
		3 & $\set{g_5,g_6}$
		& --  &   2  &   2  &   2  \\
		4 & $\set{g_1}$
		& --  &  --  &\g 1  &\g 1  \\
		5 & $\set{g_2,g_3,g_4}$
		& --  &  --  &   3  &   2  \\
		6 & $\set{g_4}$
		& --  &  --  &  --  &\g 1  \\
		\hline\hline
	\end{tabular}
	}
	\label{tab:ex-algorithm-concepts}
	\vspace{-3.0mm}
\end{table}

%% file: tex/tbl-experiments.tex
\newcommand{\s}{\cellcolor[gray]{.7}}
\begin{table}[t]
	\centering
	\caption{Runtime in seconds of \AlgoName{} and \texttt{MinIntChange} for different datasets.}
	\resizebox{0.8\columnwidth}{!}
	{
	\begin{tabular}{l|ccc|ccccc}
		\hline\hline
		DS & \# Objs & \# Attrs & $\gamma$ & $\Delta$ & \# Ptrns & $\theta$ & $t_{\text{\AlgoName{}}}$ & $t_{\mathtt{MIC}}$ \\
		\hline
		EM & 61 & 9 & 0.3 & 3 & 3 & 21 & $<0.1$ & 57\\
		BK & 96 & 4 & 0.3 & 4 & 50 & 46 & $<0.1$ & 11\\
		CN & 105 & 20 & 0.8 & 2 & 5362 & 30 & 2.4 & 28\s\\
		CU & 108 & 5 & 0.3 & 5 & 4 & 27 & $<0.1$ & 1.5\\
		FF & 125 & 3 & 0.3 & 6 & 3 & 48 & $<0.1$ & 1\\
		AP & 135 & 4 & 0.01 & 5 & 1 & 19 & $<0.1$ & 34\\
		EL & 211 & 12 & 0.3 & 6 & 33 & 83 & $<0.1$ & 34\s\\
		BA & 337 & 16 & 0.5 & 4 & 736 & 91 & 1.5 & 32\s\\
		AU & 398 & 7 & 0.3 & 7 & 17 & 234 & 0.7 & 73\s\\
		HO & 506 & 13 & 0.8 & 10 & 1 & 340 & 0.7 & 57\s\\
		QU & 2178 & 25 & 0.3 & 40 & 1 & 659 & 1.3 & 28\s\\
		AB & 4177 & 8 & 0.3 & 46 & 3 & 1400 & 11 & 86\s\\
		CA & 8192 & 21 & 0.3 & 85 & 6 & 2568 & 112 & 24\s\\
		PT & 9065 & 48 & 0.3 & 2 & 1 & 2 & 45 & 14\s\\
		\hline\hline
	\end{tabular}
	}
	\label{tbl:experiments}
\end{table}